\def\year{2019}\relax
\documentclass[letterpaper]{article} 
\usepackage{aaai19}  
\usepackage{times}  
\usepackage{helvet}  
\usepackage{courier}  
\usepackage{url}  
\usepackage{graphicx}  
\usepackage{booktabs}
\usepackage{enumitem}
\usepackage{xspace}
\usepackage{multirow}
\usepackage{bm}
\usepackage{amsmath}
\usepackage{amsfonts}
\usepackage{amsthm}
\usepackage{todonotes}
\usepackage{lipsum}
\usepackage{subcaption}
\DeclareMathOperator*{\argmin}{\arg\!\min}
\begin{document}
%

\title{Fast and Simple Mixture of Softmaxes\\ with BPE and Hybrid-LightRNN for Language Generation}
\author{Xiang Kong, Qizhe Xie, Zihang Dai, Eduard Hovy\\
  Language Technologies Institute\\
  Carnegie Mellon University\\
  \{\tt xiangk, qizhex, dzihang, hovy\}@cs.cmu.edu
}
\maketitle
\begin{abstract}
Mixture of Softmaxes (MoS) has been shown to be effective at addressing the expressiveness limitation of Softmax-based models. 
Despite the known advantage, MoS is practically sealed by its large consumption of memory and computational time due to the need of computing multiple Softmaxes.
In this work, we set out to unleash the power of MoS in practical applications by investigating improved word coding schemes, which could effectively reduce the vocabulary size and hence relieve the memory and computation burden. 
We show both BPE and our proposed Hybrid-LightRNN lead to improved encoding mechanisms that can halve the time and memory consumption of MoS without performance losses. 
With MoS, we achieve an improvement of $1.5$ BLEU scores on IWSLT 2014 German-to-English corpus and an improvement of $0.76$ CIDEr score on image captioning. 
Moreover, on the larger WMT 2014 machine translation dataset, our MoS-boosted Transformer yields $29.6$ BLEU score for English-to-German and $42.1$ BLEU score for English-to-French, outperforming the single-Softmax Transformer by $0.9$ and $0.4$ BLEU scores respectively and achieving the state-of-the-art result on WMT 2014 English-to-German task.
\end{abstract}
\section{Introduction}
\label{sec:intro}
Sequence-to-Sequence model (seq2seq) \cite{sutskever2014sequence,bahdanau2014neural} has led to significant research progress on language generation over the last few years. A typical seq2seq model employs an auto-regressive factorization of the joint distribution and outputs the conditional probability of each token given the previous tokens. A standard approach to calculate the conditional probability is to apply the Softmax function over the logits. 

Though seq2seq models with a standard Softmax output function are largely effective, \citeauthor{yang2017breaking}~\shortcite{yang2017breaking} show that the standard Softmax formulation limits the expressiveness of the generation model and results in the Softmax bottleneck.
They propose Mixture of Softmaxes (MoS) to address this issue and demonstrate improved performances on language modeling. 
However, MoS poses a non-negligible burden on the computation time and the memory consumption. Specifically, MoS outputs a weighted average of $K$ Softmax components, where computing each Softmax involves a huge dot-product between the hidden state and the embedding matrix, costing a considerable amount of time and memory.

To address the aforementioned drawbacks, a natural idea is to improve the time- and memory-efficiency of computing each Softmax.
On a high level, we aim at an encoding mechanism of the vocabulary so that each word can be represented as a code sequence.
Then, computing a single Softmax reduces to the product of a sequence of conditional code distributions.
Given a code dictionary size, the number of possible words that can be represented increases \textit{exponentially} w.r.t. the code sequence length, while the computation and memory cost 
only increases \textit{linearly}.
Hence, such an encoding scheme can theoretically reduce the time and memory consumption exponentially. 
Clearly, some encoding schemes must have better statistical properties than others and thus lead to better empirical performances.
Ideally, the encoding could be learned directly from the data. 

In this work, we investigate two algorithms for these purposes: 
The first one is called Hybrid-LightRNN, which \textit{learns} a encoding mechanism from the data based on the language modeling objective.
The other one is Byte Pair Encoding (BPE) \cite{gage1994new,sennrich2016neural}, which was originally proposed to help with translating rare words. 
When evaluated on machine translation (MT) and image captioning, both of these approaches can effectively reduce the time and memory consumption of MoS with no performance losses. 
Specifically, utilizing MoS brings a performance gain of up to $1.5$ BLEU scores on IWSLT 2014 German to English and $0.76$ CIDEr scores on image captioning. 
On WMT 2014 machine translation benchmarks, we achieve a BLEU score of $29.6$ on English-to-German and $42.1$ on English-to-French, leading to a state-of-the-art result on the WMT 2014 English-to-German task.

Our contribution is two-fold. Firstly, we propose to use Hybrid-LightRNN and BPE to make MoS time- and memory-efficient. Secondly, we demonstrate the empirical effectiveness of MoS on sentence generation by improved results on machine translation and image captioning. 

\section{Background: Mixture of Softmaxes}
Mixture of Softmaxes (MoS)~\cite{yang2017breaking} is introduced to address the expressiveness limitations of Softmax-based models. In this section, we briefly review the motivation and the formulation of MoS. 

With the autoregressive factorization, a generation model estimates the distribution of the next token $x$ given the context $c$. In language modeling, the context is composed of previous words of $x$. In conditional generation tasks such as MT or image captioning, the context also contains the source sentence or the image. Let $P^*(X \mid c_i)$ denote the ground-truth distribution of the next token given context $c$. Then the standard Softmax function computes the probability distribution $P_{\theta}(x \mid c)$ as 
\begin{equation*}
	P_{\theta}(x \mid c) = \frac{\textup{exp}\:\mathbf{h}_{c}^{\top}\mathbf{w}_{x}}{\sum_{x^{'}}\textup{exp}\:\mathbf{h}_{c}^{\top}\mathbf{w}_{x^{'}}}
\end{equation*}

where $\mathbf{h}_{c}$ is the context vector or the RNN hidden state and $\mathbf{w}_x$ is the word embedding.

\paragraph{Softmax Bottleneck}\citeauthor{yang2017breaking}~\shortcite{yang2017breaking} show the expressiveness limitation of the Softmax function from a matrix factorization perspective. 
Specifically, suppose that the number of valid contexts is finite. We list all contexts as $c_1, c_2, \cdots, c_N$. 
Let $\mathbf{A}\in \mathbb{R}^{N\times V}, \mathbf{W}\in \mathbb{R}^{V\times d}, \mathbf{H}\in \mathbb{R}^{N\times d}$ denote the log probability of the ground-truth distribution, the word embedding matrix and the context representation matrix respectively, where $N$ is the number of contexts, $V$ is the vocabulary size and $d$ is the dimensionality of the embedding vector and the context vector. 
In other words, $\mathbf{A}_{i,j}=\log P^*(x_j \mid c_i), \mathbf{W}_j=\mathbf{w}_{x_j}, \mathbf{H}_i=\mathbf{h}_{c_i}$. 

Let $F(\mathbf{A})$ denote all matrices obtained by applying row-wise shifting to $\mathbf{A}$.
Since all matrices in $F(\mathbf{A})$ result in the same probability distribution due to the normalization term in the Softmax, the Softmax function can output the ground-truth distribution $P^*$ if and only if the factorization $\mathbf{H}\mathbf{W}^{\top}$ approximate any matrix in $F(\mathbf{A})$.

However, in language generation tasks, matrices in $F(\mathbf{A})$ cannot be approximated by $\mathbf{H}\mathbf{W}^{\top}$ because of the differences in their matrix ranks. More specifically, the rank of $\mathbf{H}\mathbf{W}^{\top}$ is limited by the embedding vector dimensionality $d$. In comparison, as shown in ~\citeauthor{yang2017breaking}~\shortcite{yang2017breaking}, $\mathbf{A}$ and any other matrices within $F(\mathbf{A})$ have similar high ranks since different contexts result in highly different probability distributions of the next token. Consequently, the ground-truth distribution $P^*$ cannot be approximated by the Softmax distribution $P_{\theta}$, which results in the Softmax Bottleneck. 

\paragraph{MoS}
To tackle the Softmax bottleneck problem, MoS formulate the distribution as the weighted average of $K$ Softmax components:
\begin{equation}
\begin{split}
P_{\theta}(x \mid c)=\sum_{k=1}^{K}\pi_{c,k}\frac{\textup{exp}\:\mathbf{h}_{c,k}^{\top}\mathbf{w}_{x}}{\sum_{x^{'}}\textup{exp}\:\mathbf{h}_{c,k}^{\top}\mathbf{w}_{x^{'}}}
\end{split}
\label{eq:mos}
\end{equation}
where $\pi_{c,k}$ is the mixture weight of the $k$-th Softmax component and $\mathbf{h}_{c,k}$ is the $k$-th context vector. On language modeling, it has been shown empirically that such a formulation leads to a high rank matrix. Note that since all Softmaxes share the same word embedding matrix, the number of parameters do not increase rapidly with more mixtures, preventing overfitting.

The mixture weight and the context vectors are computed as 
\begin{equation}
\begin{aligned}
\pi_{c,k} &= \frac{\textup{exp}\:\mathbf{g}^{\top} \mathbf{w}^{(\pi)}_{k}}{\sum_{k'=1}^K \textup{exp}\:\mathbf{g}^{\top}\mathbf{w}^{(\pi)}_{k'}} \\
\mathbf{h}_{c,k} &= \textup{tanh}(\mathbf{W}^{(h)}_{k} \mathbf{g}) \\
\end{aligned}
\end{equation}
where $\mathbf{g}$ denotes a vector representation of the context $c$. $\mathbf{w}^{(\pi)}$ and $\mathbf{W}^{(h)}$ denote the parameters of the mixture weight and the parameters of the context vector with a slight abuse of notation. 

In our machine translation experiments, the attention model~\cite{bahdanau2014neural} is employed to obtain an context vector of the source sentence. $\mathbf{g}$ is obtained by passing the concatenation of the context vector and the RNN hidden state through an MLP.  
In the captioning case, the decoder is a vanilla RNN and the vector representation $\mathbf{g}$ is the decoder's hidden state. 

\paragraph{Time and Memory Cost} As shown in Eqn. \ref{eq:mos}, MoS computes $K$ Softmaxes and output the weighted average of the $K$ probability distributions. Though MoS effectively increases the expressiveness of a generation model, it also incurs a large time and memory cost since it needs to perform $K$ Softmax operations on the whole vocabulary. The time and memory costs not only hinder rapid algorithm developments but also limit the mixture number when resources are limited, restricting the power of MoS.

\section{Encoding Words for Efficient MoS}
In this section, we introduce two word encoding algorithms to reduce the memory and time consumptions of MoS. 
We aim to obtain an encoding mechanism of each word where the number of potential codes is much smaller than the vocabulary size. In theory, given a code dictionary, the number of possible words that can be represented increases exponentially w.r.t. the code sequence length, while the computation cost only increases linearly.
Then a generation model is trained to output a code sequence to generate a sentence. By decomposing words into shared codewords, the Softmax in the generation model only needs to be computed over the code dictionary. However, the encoding function need to be optimized to reflect semantic correlations between words since the semantic representations of words are shared through the embeddings of the codes. 

\subsection{Background: Learning Encoding Mechanisms Using Optimal Transport}
We first provide an optimal transport (OT) \cite{peyre2017computational} perspective of learning the encoding mechanism. Broadly speaking, optimal transport is the assignment problem between probability distributions. In the case of learning encoding mechanisms, the probability distributions are simply the delta distribution for each word and each code sequence. We define the following Wasserstein distance between the word space and the code sequence space. 

\begin{equation}
\begin{aligned}
	\min_{T}\ &\sum_{1 \leq i \leq |V|,\ \bm{s} \in S} C_{i, \bm{s}} T_{i, \bm{s}}\\
	\mathrm{s.t.}\ &T_{i, \bm{s}} \in \{0,1\} \\
	& \sum_i T_{i, \bm{s}}=1,\ \sum_j T_{i, \bm{s}}=1 \\
\end{aligned}
\label{eq:ot}
\end{equation}

where $i$ enumerates words in the vocabulary $V$ and $\bm{S}$ is the set of all code sequences. 
$T_{i, \bm{s}}$ is an indicator function of whether word $\bar{x}_i$ is assigned to code sequence $\bm{s}$. The constrains over $T$ ensures that each word is only mapped to a code sequence and that each code sequence is only mapped to a word. Hence a valid $T$ would naturally result in a desired bijection mapping. $C_{i, \bm{s}}$ is the cost of assigning word $\bar{x}_i$ to code sequence $\bm{s}$ and $\sum_{1 \leq i \leq V, \bm{s} \in \bm{S}} C_{i, \bm{s}} T_{i, \bm{s}}$ is the overall cost and the optimization objective.
For simplicity, we assume that the number of possible code sequences $|\bm{S}|$ is equal to the vocabulary size, since we can always add unused tokens to the vocabulary and assign them to redundant code sequences. 

\subsection{Hybrid-LightRNN}
As mentioned earlier, the encoding function should be learned so that words can effectively share semantics through common codes. More importantly, since the encoding function is used in language generation tasks, it is desirable to have encoded sequences that are easy to model by current RNN-based models. As language modeling can be used to measure the difficulty of modeling the code sequences, we propose Hybrid-LightRNN that optimizes the encoding function according to the language modeling objective.  

To compute the probability of a sentence under the current encoding function, we replace each sentence $\bm{x}$ by its code sequence and compute the log probability of its corresponding code sequence. Formally, let $g(\bar{x})=\bm{s}=(s_1, s_2, \cdots , s_{M})$ be the encoding function which maps a word $\bar{x}$ into a code sequence $\bm{s}$. The log probability of a sentence is as follows when an encoding function is employed:

\begin{equation}
\begin{aligned}
\log P_{\Theta}(\bm{x})&=\sum_{k} \log P_{\Theta}(x_k \mid x_1, \cdots, x_{k-1}) \\
&= \sum_{k} \sum_{j} \log P_{\Theta}(g(x_i)_j \mid Z_{k,j}) \\
\end{aligned}
\label{eq:log-likelihood}
\end{equation}

\noindent where $Z_{k,j}=g(x_1)_1, \cdots, g(x_1)_M, \cdots,  g(x_k)_1, \cdots, g(x_k)_{j-1}
$ is the concatenation of code sequences of the context and $\Theta$ is the parameters of a neural language model. Then the optimal encoding function is defined as:
\begin{equation}
\argmin_g\min_{\Theta} \sum_{\bm{x} \in X} -\log P_{\Theta}(\bm{x})
\label{eq:lm_obj}
\end{equation}
where $X$ is the training corpus. 

\paragraph{Optimization} Ideally, for each encoding function $g(\cdot)$, we would like to find the optimal language modeling cost. However, it is too computationally heavy to enumerate the combinatorial possibilities of encoding function and evaluate the language modeling performance. 
Instead, we would like to jointly optimize the encoding function $g(\cdot)$ and the language model parameters $\Theta$. However, the encoding function  $g(\cdot)$ is represented by discrete parameters, hence we resort to an approximated algorithm.
The high-level idea of the approximated algorithm is to iteratively optimize one of the language model parameters $\Theta$ and the encoding function $g(\cdot)$ while keeping the other one fixed. Since all language model parameters in $\Theta$ are fully differentiable, we can simply utilize SGD to optimize them. 
Then, the core difficulty lies in the step of optimizing the discrete parameters of $g(\cdot)$, during which, ideally, we want the following two properties to hold
\begin{itemize}
\item The encoding function remains valid. In other words, the mapping between words and code sequences remains bijections. 
\item The language modeling objective function is decreased. 
\end{itemize}

At first glance, this optimization problem seems intractable since there are combinatorially many possible $g(\cdot)$. 
However, since finding the optimal mapping is naturally an assignment problem, we can rely on existing algorithms of optimal transport if we can approximate the language modeling loss function by the Wasserstein distance defined in Eqn. \ref{eq:ot}. 
The key idea here is to decompose the corpus level likelihood to the encoding decisions of each word. More specifically, in the language modeling objective, for each word, we are measuring the likelihood of its current code sequence for each occurrence in the training data. Naturally, we can define the cost of assigning the word to the corresponding code sequences by the likelihood of other code sequences.

Formally, the cost of assigning word $\bar{x}_i$ to code sequence $\bm{s}$ can be defined as $\sum_{\bm{x} \in X} \sum_{k} \mathbb{I}(\bar{x}_i=x_k) \log -P(\bm{s} \mid Z_{k,1})$ where $\mathbb{I}(\cdot)$ is the indicator function. Here, since the context is encoded by the original encoding function, we implicitly assume the independence between the costs of  different words' mapping.
We further assume the independence between codes and approximate $\log -P(\bm{s} \mid Z_{k,1})$ as $\sum_j \log -P(\bm{s}_j \mid Z_{k,j})$ to avoid evaluating the language model for $|\bm{S}|$ times. Finally, we obtain the cost function as follows: 

\begin{equation}
\begin{aligned}
C_{i,\bm{s}}&= \sum_{\bm{x} \in X} \sum_{k} \mathbb{I}(\bar{x}_i=x_k) \sum_{j} \log -P(s_j \mid Z_{k,j})
\end{aligned}
\label{eq:likelihood}
\end{equation}

Note that, when we use the original encoding function, i.e., $T_{i,\bm{s}}=\mathbb{I}(g(\bar{x}_i)=\bm{s})$, the optimal transport objective equals to the current language modeling likelihood, i.e., $\sum_{i, \bm{s}}C_{i,\bm{s}}T_{i,\bm{s}}=\sum_{x\in X} -\log P(\bm{x})$. Hence the  Wasserstein distance will always be lower than the current language modeling cost. However, because of the independence assumptions, the language modeling loss is not guaranteed to decrease after optimizing the encoding function. 

A canonical solution to the optimal transport problem is the minimum cost maximum flow (MCMF) algorithm \cite{ahuja1993network}. 
However, the computation complexity of the MCMF is $O(|V|^3)$. 
Following LightRNN \cite{li2016lightrnn}, we adopt an $\frac{1}{2}$-approximation algorithm \cite{preis1999linear}, which has a complexity of $O(|V|^2)$. 
With the approximation algorithm, the time consumption of solving the optimal transport problem only constitutes a small proportion of the whole LightRNN algorithm when taking the time of training the neural language model into account.

\paragraph{Increasing the Capacity for Frequent Words}
Although the algorithm does not require the maximum code sequence length $M$ to be small, in our experiments, we set $M$ to $2$ since the dictionary size can already be reduced to $O(\sqrt{|V|})$ if the first code and the second code can take $\sqrt{|V|}$ values respectively. 
However, if we only uses $O(\sqrt{|V|})$ number of codes to model all words in the vocabulary, though the efficiency is improved greatly, the capacity of the model is hurt significantly, since each word is forced to share embeddings with $2 \times \sqrt{|V|} - 1$ words which share the first code or the second code with it. 

As a result, the encoding function should assign exclusive codes to important words. Since frequent words have a large impact on the overall performance, we set the encoding function so that the codes of the most frequent $K$ words are not shared with other words. Specifically, for word $\bar{x}_i\ (i < K)$, we manually specify their code sequences to be a length $1$ sequence $(i)$. For all other words, their code sequences do not contain code $i$ and are learned using the optimal transport objective.

Since the code sequence has maximum length two, we can use the following table to represent the encoding function, where the row and column denotes the first and the second respectively:
\begin{equation*}
A=\begin{bmatrix}
	D & \mathrm{UNK} \\
	\mathrm{UNK} & L \\
   \end{bmatrix}
\end{equation*}
where the matrix $D$ is a sparse diagonal matrix to which frequent words are assigned.  
$L\in \mathbb{Z}^{d_1 \times d_2}$ is a dense matrix learned through optimal transport.
To fit $V$ words into the table, the dimensions of $D$ and $L$ should satisfy $K + d_1 \times d_2 \geq |V|$.

LightRNN \cite{li2016lightrnn} is a special case of Hybrid-LightRNN where they do not  model frequent words separately. We will show in the experiments that it is very important to model frequent words separately. Furthermore, LightRNN defines the dimension of the first code to be equal to the dimension of the second code $d_1=d_2=\sqrt{V}$ for best efficiency. However, when one dimension is larger, the model can have more embedding vectors and has a larger capacity, which also results in an encoding mechanism similar to the hierarchical Softmax \cite{morin2005hierarchical}.

\subsection{Byte Pair Encoding (BPE)}
\label{sec:bpe}
Byte Pair Encoding~(BPE) \cite{gage1994new,sennrich2016neural} was introduced to address the difficulties of translating rare words and out-of-vocabulary words in machine translation. BPE is of interest here since it can reduce the vocabulary size effectively and can speedup the computation of Softmaxes.

In the encoding learned by the BPE, each code is a subword. 
Formally, BPE learns the code dictionary $\bm{S}$ as follows: We initialize the code dictionary as the set of all possible characters and break all words into sequences of codes. Then we iteratively run the following steps to add new codes to the dictionary:
\begin{itemize}
	\item[1.] Count the frequency of all code pairs within training data. Find out the most frequent pair/bigram of codes $A$ and $B$.
	\item[2.] Add the new code $AB$ to the dictionary. Replace all occurrence of pair $(A, B)$ with $AB$.
	\item[3.] End the iteration if the dictionary size reaches a threshold. Otherwise go to step $1$.
\end{itemize}

BPE is an algorithm based on heuristics. However, the strong inductive bias of BPE always gives more capacity to frequent words when it comes to the tradeoff between efficiency and capacity if we vary the subword unit dictionary size, since the more frequent words will be segmented into fewer parts, which will lead to more exclusive embeddings instead of shared embeddings. 

When we use a larger code dictionary, more frequent words and subwords are added to the dictionary and their semantics are modeled by separate embedding vectors, leading to a larger model capacity. On the other hand, the model efficiency is improved with a smaller subword dictionary.

\section{Related Work}
Apart from the previously mentioned related works, mixture of Softmaxes is closely related to works that mix  representation vectors~\cite{eigen2013learning,shazeer2017outrageously}. ~\citeauthor{yang2017breaking}~\shortcite{yang2017breaking} show that this approach does not solve the softmax bottleneck problem.

Hierarchical Softmax~\cite{morin2005hierarchical} is an extensively studied technique to improve the efficiency of Softmaxes. 
\citeauthor{morin2005hierarchical}~\shortcite{morin2005hierarchical} uses the synsets in the WordNet to build the hierarchical tree. 
\citeauthor{mnih2009scalable}~\shortcite{mnih2009scalable} propose to learn the hierarchical tree with a clustering algorithm. The idea of separately modeling frequent words is also explored in Adaptive Softmax~\cite{grave2016efficient}.
Although hierarchical Softmax can reduce the time and memory consumptions during training, it still requires computing the Softmax over the whole vocabulary during testing. 
Noise Contrastive Estimation~\cite{gutmann2012noise,mnih2012fast} and Negative Sampling~\cite{mikolov2013distributed} can also speed up Softmax during training.

\section{Experiments}

\begin{table*}
\centering

\small
\begin{tabular}{c|c|c|ccc} 
\toprule
& & Machine Translation (IWSLT) & \multicolumn{3}{c}{Image Captioning (MSCOCO)}  \\
Model &\# Softmaxes & BLEU & BLEU-4 & METEOR & CIDEr \\
\midrule
Baseline & 1& 27.41 $\pm$ 0.15 & 29.64 $\pm$ 0.20 & 23.60 $\pm$ 0.12 & 88.50 $\pm$ 0.47  \\ 
Hybrid-LightRNN-MoS & 9&28.79 $\pm$ 0.23 & 30.02 $\pm$ 0.16 & 23.87 $\pm$ 0.18 & 88.96 $\pm$ 0.21 \\
BPE-MoS & 9& \textbf{28.91 $\pm$ 0.06} &\textbf{30.06 $\pm$ 0.10} & \textbf{24.00 $\pm$ 0.24}  & \textbf{89.26 $\pm$ 0.11} \\
\bottomrule
\end{tabular}
\caption{Overall performance comparisons on IWSLT and MSCOCO}
\label{tab:overall}
\end{table*}

\begin{table}[t]
\centering
\small
\begin{tabular}{l|cc}
\toprule
{\bf Model}  &   {\bf EN-DE}  & {\bf EN-FR}    \\
\midrule
\citeauthor{wu2016google}~\shortcite{wu2016google}  &  26.30 & 41.16\\
\citeauthor{shazeer2017outrageously}~\shortcite{shazeer2017outrageously} & 26.03 & 40.56 \\
\citeauthor{gehring2017convolutional}~\shortcite{gehring2017convolutional}  & 26.43 & 41.62\\
\citeauthor{vaswani2017attention}~\shortcite{vaswani2017attention} & 28.4 & 41.8\\
\citeauthor{dehghani2018universal}~\shortcite{dehghani2018universal} & 28.9 & N/A \\
\citeauthor{shaw2018self}~\shortcite{shaw2018self} & 29.2 & 41.5 \\
\citeauthor{ott2018scaling}~\shortcite{ott2018scaling} & 29.3 & {\bf 43.2} \\
\midrule
 Our Transformer (Base)  & 27.4 & 38.2 \\
 Transformer-MoS (Base) & 28.0 & 39.1\\
 \midrule
  Our Transformer (Big)    & 28.7 & 41.7 \\
  Transformer-MoS (Big) &  \bf {29.6} & 42.1\\
\bottomrule         
\end{tabular}
\caption{Experiment results on the WMT 2014 English-German (EN-DE) and English-French (EN-FR) where Transformer-MoS denotes the Transformer model with MoS. }
\label{tab:wmt_res} 
\end{table}

\begin{table*}
\centering
\small
\begin{tabular}{c|c|ccc|ccc}
\toprule
\multirow{2}{*}{\# Softmaxes} & \multirow{2}{*}{Model} &  \multicolumn{3}{|c|}{Machine Translation (IWSLT)} &\multicolumn{3}{c}{Image Captioning (MSCOCO)} \\
&  & Memory  & Speed  &  BLEU  &  Memory   & Speed & BLEU-4 \\
\midrule
\multirow{3}{*}{1} & Baseline  & 5.18 & 40.38 & 27.41 $\pm$ 0.15 &  0.96 & 13.29 & 29.64 $\pm$ 0.20 \\
 & Hybrid-LightRNN-Baseline &  3.61 & 33.57 & 27.43 $\pm$ 0.21  & 0.73 & 10.35 & 29.69 $\pm$ 0.15 \\
 & BPE-Baseline & \textbf{3.52} & \textbf{32.78} &  27.37 $\pm$ 0.17 &\textbf{0.70} & \textbf{9.96} &29.68 $\pm$ 0.12 \\
\midrule
\multirow{3}{*}{3} & MoS  & 10.24 & 89.94 & 28.42 $\pm$ 0.14  &  1.39 & 25.66 & 29.91 $\pm$ 0.14\\
 & Hybrid-LightRNN-MoS  & 5.37 & 49.19 & 28.36 $\pm$ 0.11  & 1.01 & 15.49 & 29.93 $\pm$ 0.14\\
 & BPE-MoS  & \textbf{5.27} & \textbf{46.83} & 28.47 $\pm$ 0.16 & \textbf{0.99} & \textbf{15.03} & 29.96 $\pm$ 0.08  \\
\bottomrule
\end{tabular}
\caption{Memory and time efficiency comparisons on MT and image captioning when using the same number of Softmaxes. Bold faces highlight the best in the corresponding category. The shown memory is in GB and the speed is in ms/batch.}
\label{tab:same-soft}
\end{table*}

\begin{table*}[ht]
\centering

\small
  \begin{tabular}{cc|cc|c} 
 \toprule
 
Memory (GB) & Speed (ms/batch)&  Model  & \# Mixtures &  BLEU \\  
\midrule
\multirow{3}{*}{5.3} & \multirow{3}{*}{45.5} & Baseline &   $1 \times 30\mathrm{k}$ & 27.41 $\pm$  0.15 \\
  &  & Hybrid-LightRNN-MoS   & $3 \times 10\mathrm{k}$   & 28.36 $\pm$ 0.11  \\
  &  & BPE-MoS  & $3\times 10\mathrm{k}$  & \textbf{28.47 $\pm$ 0.16}  \\
 \midrule
\multirow{3}{*}{10.4}& \multirow{3}{*}{90.4} &  MoS  &  $3 \times 30\mathrm{k}$   & 28.42 $\pm$ 0.14  \\
  &  &  Hybrid-LightRNN-MoS  &  $9 \times 10\mathrm{k}$  & 28.79 $\pm$ 0.23 \\
  &  &  BPE-MoS  & $9 \times 10 \mathrm{k}$ &  \textbf{28.91 $\pm$ 0.06}  \\
 \bottomrule
\end{tabular}
 \caption{Comparisons on IWSLT under the same memory and time budget. The Softmaxes size is number of Softmaxes $\times$ Softmax dictionary size. Bold faces highlight the best in the corresponding category}
 \label{tab:nmt-results}
\end{table*}

\begin{table*}[ht]
\centering

\small
  \begin{tabular}{cc|cc|ccc} 
 \toprule
 
Memory (GB) & Speed (ms/batch)&  Model  & Softmaxes Size &  BLEU-4 & METEOR & CIDEr \\  
\midrule
\multirow{3}{*}{1.0} & \multirow{3}{*}{14.6} & MoS &   $1 \times 10\mathrm{k}$ & 29.64 $\pm$ 0.20 & 23.60 $\pm$ 0.12 & 88.50 $\pm$ 0.47 \\
  &  & Hybrid-LightRNN-MoS   & $3 \times 3\mathrm{k}$  & 29.93 $\pm$ 0.14 &\textbf{23.74 $\pm$ 0.29} & 88.52 $\pm$ 0.18  \\
  &  & BPE-MoS  & $3\times 3\mathrm{k}$  & \textbf{29.96 $\pm$ 0.08} & 23.67 $\pm$ 0.34 & \textbf{88.61 $\pm$ 0.27}  \\
 \midrule
\multirow{3}{*}{4.3}& \multirow{3}{*}{26.4} &  MoS  &   $3 \times 10\mathrm{k}$   & 29.91 $\pm$ 0.14 & 23.69 $\pm$ 0.20 & 88.84 $\pm $0.66   \\
  &  &  Hybrid-LightRNN-MoS  &   $9\times 3\mathrm{k}$  & 30.02 $\pm$ 0.16 & 23.87 $\pm$ 0.18 & 88.96 $\pm$ 0.21 \\
  &  &  BPE-MoS  & $9 \times 3\mathrm{k}$  & \textbf{30.06 $\pm$ 0.10} & \textbf{24.00 $\pm$ 0.24}  & \textbf{89.26 $\pm$ 0.11}  \\
 \bottomrule
\end{tabular}
 \caption{Comparisons on image captioning under the same memory and time budget. The Softmaxes size is number of Softmaxes $\times$ Softmax dictionary size.  Bold faces highlight the best ones in each category.}
 \label{tab:ic-results}
\end{table*}

\begin{table*}[ht]
\centering

\small
 \begin{tabular}{c|ccccccccc} 
 \toprule
 row & \multicolumn{8}{c}{words} \\
 \midrule
  45 & 700 & 3.3 & 28 & 19 & 7 & 86 & 35 & ... \\
 48 & around & between & by & into & down & for & off & ...\\
 54 & mined & imaged & advised & pickled & outfitted & filled & withheld & ...\\
 91 & bristol & chinatown & rochester &  kingston & guangdong & guangzhou & chongqing & ...\\
 93 & pursuing & posing & proposing &reacting & replacing & blogging &pointing & ...\\
  \bottomrule
 \end{tabular}
 \caption{Example mapping table where the row denotes the first code and the column denotes the second code. Numbers and places are grouped together in row 45, 91. Syntactically similar words are also grouped together in row 48, 54 and 93. }
 \label{tab:allo-table}
\end{table*}

In this section, we describe our experiments on machine translation and image captioning and study our models quantitatively and qualitatively. 
\subsection{Experiment Settings}
\paragraph{Machine Translation}
We first evaluate our models on the IWSLT 2014 German to English~(DE-EN) dataset ~\cite{cettolo2013report}. 

We employ an LSTM~\cite{hochreiter1997long} seq2seq model with the dot-product attention~\cite{bahdanau2014neural,luong2015effective} as the baseline. We build the baseline using the PyTorch code from \citeauthor{dai2018credit}~\shortcite{dai2018credit}. For Hybrid-LightRNN, we set $K$ to $9,652$ and set $d_1$ and $d_2$ to $174$ to represent a total of $30\mathrm{K}$ words. As model performances exhibit small variances on IWSLT, we run each experiment for five times with different random seeds and report the average performance and the standard deviation.

We also test our best model on the standard WMT 2014 English-to-German~(EN-DE) and English-to-French~(EN-FR) benchmarks, consisting of $4.5\mathrm{M}$ and $36\mathrm{M}$ sentence pairs respectively. We follow the preprocessing steps of ConvS2S \cite{gehring2017convolutional} for the EN-FR task.
We employ BPE with $32\mathrm{K}$ merge operations for both tasks.  The Transformer model~\cite{vaswani2017attention} is employed as our baseline. Our configuration largely follows the configuration of ~\citeauthor{vaswani2017attention}~\shortcite{vaswani2017attention}, except that we multiply the original learning rate by 0.8 for the Transformer equipped with MoS.  Specifically, we test both the Base configuration and Big configuration, which respectively have embeddings of dimension 512 and 1024, the dimension of the inner layer 2048 and 4096 and the number of attention heads 8 and 16. We used the Adam optimizer~\cite{kingma2014adam} with $\beta_{1}=0.9$, $\beta_{2}=0.98$, and  $\epsilon=10^{-9}$.
We set the mixture number to $9$. We use the corpus-level BLEU score~\cite{papineni2002bleu} as the evaluation metric. Our Transformer training and evaluation code is based on an open source toolkit THUMT~\cite{zhang2017thumt}.

\paragraph{Image Captioning} We conduct experiments on the MSCOCO dataset~\cite{lin2014microsoft} and follow the same preprocessing procedure and the train/validation/test split as used in \citeauthor{karpathy2015deep}~\shortcite{karpathy2015deep}.
We use the Neural Image Caption (NIC) model~\cite{vinyals2015show} as the baseline model. Following ~\citeauthor{dai2018credit}~\shortcite{dai2018credit} and \citeauthor{kong2018neural}~\shortcite{dai2018credit}, we employ a pretrained 101-layer ResNet~\cite{he2016deep} instead of a GoogLeNet to extract a feature vector from an input image. We employ an LSTM of size $512$ as the decoder.
We report BLEU-4, METEOR and CIDERr scores using the scripts provided by \citeauthor{chen2015microsoft}~\shortcite{chen2015microsoft}.

\paragraph{Experiment Details}
For the BPE and Hybrid-LightRNN, we set the code dictionary sizes to $10\mathrm{K}$ for IWSLT and $3\mathrm{K}$ for MSCOCO. 
We measure the speed and memory usage on a Titan X with PyTorch version v0.3.1 and CUDA 9.0. 

\begin{table}[ht]
\centering
\small

 \begin{tabular}{l|c|c|c} 
 \toprule
Mapping Table & Table Size & Learned Table & BLEU \\
 \midrule
  Hybrid-LightRNN & $10\mathrm{k}$ & \multirow{3}{*}{Yes} & 30.07 \\ 
 Hybrid-LightRNN &  $5\mathrm{k}$ & & 29.69\\ 
 Hybrid-LightRNN & $1\mathrm{k}$ & & 28.73  \\ 
 \midrule
 LightRNN & $0.2\mathrm{k}$ & Yes & 27.39 \\ 
 \midrule
 Frequency table & $0.2\mathrm{k}$ & \multirow{2}{*}{No} & 25.84 \\ 
 Random table & $0.2\mathrm{k}$ & & 24.98 \\ 
  \bottomrule
 \end{tabular}
 \caption{Average validation BLEU on IWSLT of Hybrid-LightRNN-MoS using different mapping tables.}
 \label{tab:lightas-results}
\end{table}

\begin{table}
\centering

\small
\begin{tabular}{c|c|c} 
\toprule
Model  & OOV Translation & BLEU \\
\midrule
BPE-MoS  & Yes & 30.19\\
BPE-MoS  & No & 30.14\\
Hybrid-LightRNN-MoS & No & 30.07\\
\bottomrule
\end{tabular}
\caption{Ablation study on the importance of translating OOV words. The BLEU score are evaluated on the validation set of IWSLT.}
\label{tab:oov}

\end{table}

\subsection{Main Results}
In our experiments, we denote Hybrid-LightRNN-MoS and BPE-MoS as the seq2seq models with MoS which employ Hybrid-LightRNN and BPE respectively. The baseline seq2seq model without MoS is denoted as Baseline. 
\paragraph{Overall Performances on IWSLT and MSCOCO} We show the comparison between a standard LSTM seq2seq model with the Hybrid-LightRNN-MoS and BPE-MoS in Tab. \ref{tab:overall}. Hybrid-LightRNN-MoS and BPE-MoS both outperform the baseline on both tasks. Specifically, on machine translation, BPE-MoS can outperform the baseline by a BLEU score of $1.5$. On image captioning, BPE-MoS outperforms the baseline by $0.42$, $0.4$ and $0.76$ in terms of BLEU-4, METEOR and CIDEr respectively. 

This experiment shows that MoS can effectively improve the expressiveness of generation models by learning a high-rank log probability matrix.
As expected, the improvement is larger on MT than on image captioning, which can be explained by the differences of language complexities used in these two tasks. 
Specifically, on image captioning, the captions largely share similar patterns, resulting in a lower-rank probability matrix and a smaller improvement space. 

\paragraph{Performances on WMT 14 EN-DE and EN-FR} Since BPE is better than Hybrid-LightRNN with a small margin on IWSLT, we only test BPE-MoS on WMT. As shown in Tab. \ref{tab:wmt_res}, we achieve $29.6$ and $42.1$ BLEU scores respectively on WMT 14 EN-DE and EN-FR, improving the Transformer model by $0.9$ and $0.4$ BLEU scores. We achieve the state-of-the-art result that does not employ data augmentation on WMT 14 EN-DE. Note that data augmentation can also effectively improve the machine translation performance~\cite{edunov2018understanding}.

\paragraph{Memory and Time Efficiency} We study the memory consumption and efficiency of Hybrid-LightRNN-MoS and BPE-MoS.
As shown in Tab. \ref{tab:same-soft}, when applying on the Baseline model and the MoS model, BPE and Hybrid-LightRNN can reduce the time and memory usage with no performance losses. In addition, on MT where the vocabulary is large, they can halve the time and memory consumption when applied on MoS with $3$ mixtures. When there are more mixutres, the improvements will continue to grow since computing Softmaxes take a larger proportion of time. 

\paragraph{Comparisons under the Same Computation Budget} When computational resources are limited, BPE and LightRNN enable the use of more Softmaxes, leading to potentially higher rank probability matrices. Hence, we study the performances of BPE-MoS, Hybrid-LightRNN-MoS and MoS given the same computation budget. As shown in Tab. \ref{tab:nmt-results}. 
BPE-MoS and Hybrid-LightRNN-MoS consistently outperform the baseline and the MoS model. 

\subsection{Analysis}
In this section, we perform extensive studies to better understand our models.

\begin{figure}[t]
	\centering
	\includegraphics[width=0.4\textwidth]{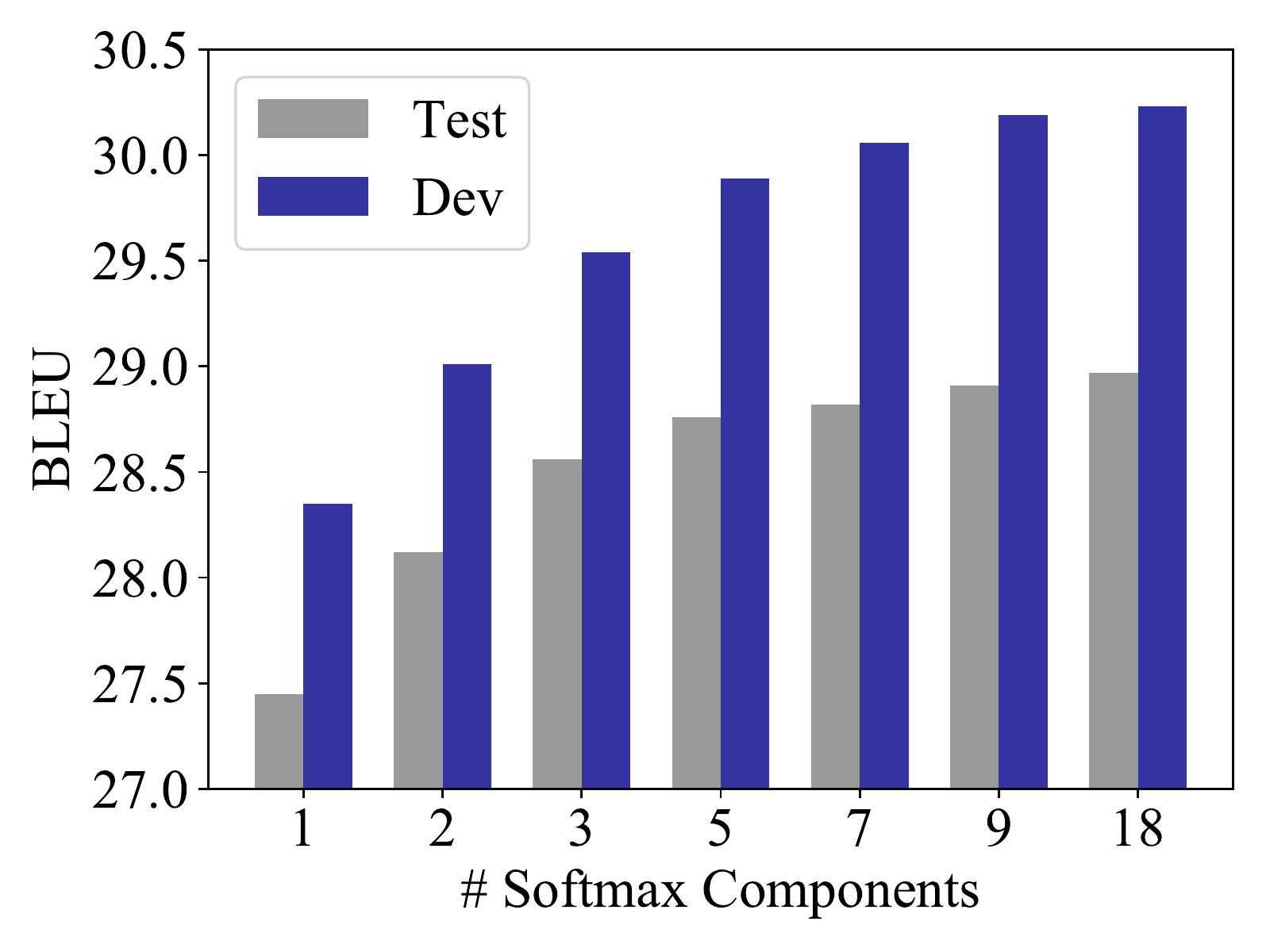}
	\caption{BPE-MoS's average validation performance over multiple runs on IWSLT with various numbers of mixture components.}
	\label{fig:bpe-as}
\end{figure}

\paragraph{Number of Softmaxes} 
Since a larger mixture number would likely to lead to a higher rank log probability matrix, we verify whether a larger mixture number leads to a better performance.
We vary the number of mixture  in the BPE-MoS model and compare their performances on MT. 
As shown in Fig.~\ref{fig:bpe-as}, more Softmax components clearly lead to better performances. 
However, the improvement margin exhibits a diminishing return effect, which means that several Softmaxes are enough to learn a high-rank matrix.

\paragraph{Hybrid-LightRNN Ablation Study}
We further study the importance of the learned table and the importance of the model's capacity in Hybrid-LightRNN.
Firstly, we vary the dictionary size to investigate whether it is necessary to give enough capacity to frequent words. 
   
As shown in Tab. \ref{tab:lightas-results}, larger dictionary sizes consistently lead to better performances. 
Secondly, when compared with LightRNN, Hybrid-LightRNN achieves an improvement of $2.68$ BLEU score, which shows that it is necessary to employ extra capacities for frequent words.
Thirdly, as a sanity check of whether the table learning is necessary, we compare the table learned by LightRNN with the table obtained by simply sorting words based on their frequency and the table with random word allocations.
The table learned by LightRNN  outperforms models with the random table or the frequency-based table by BLEU scores of $2.41$ and $1.55$ respectively, which means that optimizing a language modeling objective learns an effective encoding function.

\paragraph{Is BPE-MoS better because of modeling OOVs?}
As indicated in Tab. \ref{tab:overall}, BPE-MoS is slightly better than Hybrid-Light-MoS on MT and image captioning. 
In principle, both Hybrid-LightRNN and BPE can model the semantics of all frequent words and rare words in the training set by sharing embeddings with other words. One exclusive advantage of BPE is the ability to generate out-of-vocabulary (OOV) words. A natural question to ask is ``how much performance difference  would OOVs cause?'' To investigate the importance of modeling OOVs, We take the best BPE-MoS model, replace all generated OOV words with UNK and test its performance.

The comparison is shown in Tab. \ref{tab:oov}. Removing the generated OOV words do lead to a performance decrease of $0.05$ BLEU score. 
However, when both Hybrid-LightRNN and BPE are disabled from translating OOV, BPE is still better than Hybrid-LightRNN by a gap of $0.07$ BLEU score. 
This result indicates that the encoding function learned by BPE better captures the data statistics than the encoding learned by Hybrid-LightRNN, showing that Hybrid-LightRNN has a lot of potentials for improvements.

\paragraph{Mapping Table Qualitative Study}
In Hybrid-LightRNN, words in the same column/row  share the same column/row embedding vector. 
Intuitively, it is important to group semantically-similar or syntactically-similar words into the same column/row. We examine whether the learned table have this property in Tab. \ref{tab:allo-table}. 
We find that most words within the same row are either semantically-similar or syntactically-similar to each other.

\section{Conclusions and Discussions}

In this work, we investigate two algorithms, i.e., Byte Pair Encoding and Hybrid-LightRNN, to reduce the vocabulary size so as to improve the memory- and time-efficiency of MoS.
We evaluate these two methods on machine translation and image captioning and show improved performances over the baseline system without MoS. 
Further, both of these methods effectively speed up the training process and reduce the memory consumption of MoS with no performance losses.  
We demonstrate the effectiveness of our models by improved performances on machine translation and image captioning. 

\bibliography{AAAI-KongX.6411}

\begin{thebibliography}{}

\bibitem[\protect\citeauthoryear{Ahuja \bgroup et al\mbox.\egroup
  }{1993}]{ahuja1993network}
Ahuja, R.~K.; Magnanti, T.~L.; Orlin, J.~B.; et~al.
\newblock 1993.
\newblock {\em Network flows: theory, algorithms, and applications}, volume~1.
\newblock Prentice hall Englewood Cliffs, NJ.

\bibitem[\protect\citeauthoryear{Bahdanau, Cho, and
  Bengio}{2014}]{bahdanau2014neural}
Bahdanau, D.; Cho, K.; and Bengio, Y.
\newblock 2014.
\newblock Neural machine translation by jointly learning to align and
  translate.
\newblock {\em arXiv preprint arXiv:1409.0473}.

\bibitem[\protect\citeauthoryear{Cettolo \bgroup et al\mbox.\egroup
  }{2014}]{cettolo2013report}
Cettolo, M.; Niehues, J.; St{\"u}ker, S.; Bentivogli, L.; and Federico, M.
\newblock 2014.
\newblock Report on the 11th iwslt evaluation campaign.
\newblock In {\em Proc. of IWSLT}.

\bibitem[\protect\citeauthoryear{Chen \bgroup et al\mbox.\egroup
  }{2015}]{chen2015microsoft}
Chen, X.; Fang, H.; Lin, T.-Y.; Vedantam, R.; Gupta, S.; Doll{\'a}r, P.; and
  Zitnick, C.~L.
\newblock 2015.
\newblock Microsoft coco captions: Data collection and evaluation server.
\newblock {\em arXiv preprint arXiv:1504.00325}.

\bibitem[\protect\citeauthoryear{Dai, Xie, and Hovy}{2018}]{dai2018credit}
Dai, Z.; Xie, Q.; and Hovy, E.
\newblock 2018.
\newblock From credit assignment to entropy regularization: Two new algorithms
  for neural sequence prediction.
\newblock {\em arXiv preprint arXiv:1804.10974}.

\bibitem[\protect\citeauthoryear{Dehghani \bgroup et al\mbox.\egroup
  }{2018}]{dehghani2018universal}
Dehghani, M.; Gouws, S.; Vinyals, O.; Uszkoreit, J.; and Kaiser, {\L}.
\newblock 2018.
\newblock Universal transformers.
\newblock {\em arXiv preprint arXiv:1807.03819}.

\bibitem[\protect\citeauthoryear{Edunov \bgroup et al\mbox.\egroup
  }{2018}]{edunov2018understanding}
Edunov, S.; Ott, M.; Auli, M.; and Grangier, D.
\newblock 2018.
\newblock Understanding back-translation at scale.
\newblock {\em arXiv preprint arXiv:1808.09381}.

\bibitem[\protect\citeauthoryear{Eigen, Ranzato, and
  Sutskever}{2013}]{eigen2013learning}
Eigen, D.; Ranzato, M.; and Sutskever, I.
\newblock 2013.
\newblock Learning factored representations in a deep mixture of experts.
\newblock {\em arXiv preprint arXiv:1312.4314}.

\bibitem[\protect\citeauthoryear{Gage}{1994}]{gage1994new}
Gage, P.
\newblock 1994.
\newblock A new algorithm for data compression.
\newblock {\em The C Users Journal} 12(2):23--38.

\bibitem[\protect\citeauthoryear{Gehring \bgroup et al\mbox.\egroup
  }{2017}]{gehring2017convolutional}
Gehring, J.; Auli, M.; Grangier, D.; Yarats, D.; and Dauphin, Y.~N.
\newblock 2017.
\newblock Convolutional sequence to sequence learning.
\newblock {\em arXiv preprint arXiv:1705.03122}.

\bibitem[\protect\citeauthoryear{Grave \bgroup et al\mbox.\egroup
  }{2016}]{grave2016efficient}
Grave, E.; Joulin, A.; Ciss{\'e}, M.; Grangier, D.; and J{\'e}gou, H.
\newblock 2016.
\newblock Efficient softmax approximation for gpus.
\newblock {\em arXiv preprint arXiv:1609.04309}.

\bibitem[\protect\citeauthoryear{Gutmann and
  Hyv{\"a}rinen}{2012}]{gutmann2012noise}
Gutmann, M.~U., and Hyv{\"a}rinen, A.
\newblock 2012.
\newblock Noise-contrastive estimation of unnormalized statistical models, with
  applications to natural image statistics.
\newblock {\em Journal of Machine Learning Research} 13(Feb):307--361.

\bibitem[\protect\citeauthoryear{He \bgroup et al\mbox.\egroup
  }{2016}]{he2016deep}
He, K.; Zhang, X.; Ren, S.; and Sun, J.
\newblock 2016.
\newblock Deep residual learning for image recognition.
\newblock In {\em Proc of CVPR},  770--778.

\bibitem[\protect\citeauthoryear{Hochreiter and
  Schmidhuber}{1997}]{hochreiter1997long}
Hochreiter, S., and Schmidhuber, J.
\newblock 1997.
\newblock Long short-term memory.
\newblock {\em Neural computation} 9(8):1735--1780.

\bibitem[\protect\citeauthoryear{Karpathy and Fei-Fei}{2015}]{karpathy2015deep}
Karpathy, A., and Fei-Fei, L.
\newblock 2015.
\newblock Deep visual-semantic alignments for generating image descriptions.
\newblock In {\em Proc. of CVPR},  3128--3137.

\bibitem[\protect\citeauthoryear{Kingma and Ba}{2014}]{kingma2014adam}
Kingma, D.~P., and Ba, J.
\newblock 2014.
\newblock Adam: A method for stochastic optimization.
\newblock {\em arXiv preprint arXiv:1412.6980}.

\bibitem[\protect\citeauthoryear{Kong \bgroup et al\mbox.\egroup
  }{2018}]{kong2018neural}
Kong, X.; Tu, Z.; Shi, S.; Hovy, E.; and Zhang, T.
\newblock 2018.
\newblock Neural machine translation with adequacy-oriented learning.
\newblock {\em arXiv preprint arXiv:1811.08541}.

\bibitem[\protect\citeauthoryear{Li \bgroup et al\mbox.\egroup
  }{2016}]{li2016lightrnn}
Li, X.; Qin, T.; Yang, J.; Hu, X.; and Liu, T.
\newblock 2016.
\newblock Lightrnn: Memory and computation-efficient recurrent neural networks.
\newblock In {\em NIPS},  4385--4393.

\bibitem[\protect\citeauthoryear{Lin \bgroup et al\mbox.\egroup
  }{2014}]{lin2014microsoft}
Lin, T.-Y.; Maire, M.; Belongie, S.; Hays, J.; Perona, P.; Ramanan, D.;
  Doll{\'a}r, P.; and Zitnick, C.~L.
\newblock 2014.
\newblock Microsoft coco: Common objects in context.
\newblock In {\em ECCV}.
\newblock Springer.

\bibitem[\protect\citeauthoryear{Luong, Pham, and
  Manning}{2015}]{luong2015effective}
Luong, M.-T.; Pham, H.; and Manning, C.~D.
\newblock 2015.
\newblock Effective approaches to attention-based neural machine translation.
\newblock {\em arXiv preprint arXiv:1508.04025}.

\bibitem[\protect\citeauthoryear{Mikolov \bgroup et al\mbox.\egroup
  }{2013}]{mikolov2013distributed}
Mikolov, T.; Sutskever, I.; Chen, K.; Corrado, G.~S.; and Dean, J.
\newblock 2013.
\newblock Distributed representations of words and phrases and their
  compositionality.
\newblock In {\em NIPS},  3111--3119.

\bibitem[\protect\citeauthoryear{Mnih and Hinton}{2009}]{mnih2009scalable}
Mnih, A., and Hinton, G.~E.
\newblock 2009.
\newblock A scalable hierarchical distributed language model.
\newblock In {\em NIPS},  1081--1088.

\bibitem[\protect\citeauthoryear{Mnih and Teh}{2012}]{mnih2012fast}
Mnih, A., and Teh, Y.~W.
\newblock 2012.
\newblock A fast and simple algorithm for training neural probabilistic
  language models.
\newblock {\em arXiv preprint arXiv:1206.6426}.

\bibitem[\protect\citeauthoryear{Morin and
  Bengio}{2005}]{morin2005hierarchical}
Morin, F., and Bengio, Y.
\newblock 2005.
\newblock Hierarchical probabilistic neural network language model.
\newblock In {\em Aistats}, volume~5,  246--252.
\newblock Citeseer.

\bibitem[\protect\citeauthoryear{Ott \bgroup et al\mbox.\egroup
  }{2018}]{ott2018scaling}
Ott, M.; Edunov, S.; Grangier, D.; and Auli, M.
\newblock 2018.
\newblock Scaling neural machine translation.
\newblock {\em arXiv preprint arXiv:1806.00187}.

\bibitem[\protect\citeauthoryear{Papineni \bgroup et al\mbox.\egroup
  }{2002}]{papineni2002bleu}
Papineni, K.; Roukos, S.; Ward, T.; and Zhu, W.-J.
\newblock 2002.
\newblock Bleu: a method for automatic evaluation of machine translation.
\newblock In {\em Proc. of ACL},  311--318.

\bibitem[\protect\citeauthoryear{Peyr{\'e} and
  Cuturi}{2017}]{peyre2017computational}
Peyr{\'e}, G., and Cuturi, M.
\newblock 2017.
\newblock Computational optimal transport.
\newblock Technical report.

\bibitem[\protect\citeauthoryear{Preis}{1999}]{preis1999linear}
Preis, R.
\newblock 1999.
\newblock Linear time 1/2-approximation algorithm for maximum weighted matching
  in general graphs.
\newblock In {\em STACS},  259--269.

\bibitem[\protect\citeauthoryear{Sennrich, Haddow, and
  Birch}{2016}]{sennrich2016neural}
Sennrich, R.; Haddow, B.; and Birch, A.
\newblock 2016.
\newblock Neural machine translation of rare words with subword units.
\newblock In {\em Proc. of ACL}, volume~1,  1715--1725.

\bibitem[\protect\citeauthoryear{Shaw, Uszkoreit, and
  Vaswani}{2018}]{shaw2018self}
Shaw, P.; Uszkoreit, J.; and Vaswani, A.
\newblock 2018.
\newblock Self-attention with relative position representations.
\newblock In {\em Proc. of NAACL}.

\bibitem[\protect\citeauthoryear{Shazeer \bgroup et al\mbox.\egroup
  }{2017}]{shazeer2017outrageously}
Shazeer, N.; Mirhoseini, A.; Maziarz, K.; Davis, A.; Le, Q.; Hinton, G.; and
  Dean, J.
\newblock 2017.
\newblock Outrageously large neural networks: The sparsely-gated
  mixture-of-experts layer.
\newblock {\em arXiv preprint arXiv:1701.06538}.

\bibitem[\protect\citeauthoryear{Sutskever, Vinyals, and
  Le}{2014}]{sutskever2014sequence}
Sutskever, I.; Vinyals, O.; and Le, Q.~V.
\newblock 2014.
\newblock Sequence to sequence learning with neural networks.
\newblock In {\em NIPS},  3104--3112.

\bibitem[\protect\citeauthoryear{Vaswani \bgroup et al\mbox.\egroup
  }{2017}]{vaswani2017attention}
Vaswani, A.; Shazeer, N.; Parmar, N.; Uszkoreit, J.; Jones, L.; Gomez, A.~N.;
  Kaiser, {\L}.; and Polosukhin, I.
\newblock 2017.
\newblock Attention is all you need.
\newblock In {\em NIPS},  5998--6008.

\bibitem[\protect\citeauthoryear{Vinyals \bgroup et al\mbox.\egroup
  }{2015}]{vinyals2015show}
Vinyals, O.; Toshev, A.; Bengio, S.; and Erhan, D.
\newblock 2015.
\newblock Show and tell: A neural image caption generator.
\newblock In {\em Proc. of CVPR},  3156--3164.
\newblock IEEE.

\bibitem[\protect\citeauthoryear{Wu \bgroup et al\mbox.\egroup
  }{2016}]{wu2016google}
Wu, Y.; Schuster, M.; Chen, Z.; Le, Q.~V.; Norouzi, M.; Macherey, W.; Krikun,
  M.; Cao, Y.; Gao, Q.; Macherey, K.; et~al.
\newblock 2016.
\newblock Google's neural machine translation system: Bridging the gap between
  human and machine translation.
\newblock {\em arXiv preprint arXiv:1609.08144}.

\bibitem[\protect\citeauthoryear{Yang \bgroup et al\mbox.\egroup
  }{2018}]{yang2017breaking}
Yang, Z.; Dai, Z.; Salakhutdinov, R.; and Cohen, W.~W.
\newblock 2018.
\newblock Breaking the softmax bottleneck: a high-rank rnn language model.
\newblock In {\em Proc. of ICLR}.

\bibitem[\protect\citeauthoryear{Zhang \bgroup et al\mbox.\egroup
  }{2017}]{zhang2017thumt}
Zhang, J.; Ding, Y.; Shen, S.; Cheng, Y.; Sun, M.; Luan, H.; and Liu, Y.
\newblock 2017.
\newblock Thumt: An open source toolkit for neural machine translation.
\newblock {\em arXiv preprint arXiv:1706.06415}.

\end{thebibliography}
\bibliographystyle{aaai}

\end{document}